\documentclass[letterpaper]{article} 
\usepackage{aaai24}
\usepackage{times}  
\usepackage{helvet}  
\usepackage{courier}  
\usepackage[hyphens]{url}  
\usepackage{graphicx} 
\def\UrlFont{\rm}  
\usepackage{natbib}  
\usepackage{caption} 
\usepackage{algorithm}
\usepackage{algorithmic}

\usepackage{xcolor}
\usepackage{amssymb}
\usepackage{bbold}
\usepackage{amsmath}
\usepackage{multirow}
\usepackage{booktabs}

\usepackage{newfloat}
\usepackage{listings}
\DeclareCaptionStyle{ruled}{labelfont=normalfont,labelsep=colon,strut=off} 
\floatstyle{ruled}
\newfloat{listing}{tb}{lst}{}
\floatname{listing}{Listing}
\title{On the Concept Trustworthiness in Concept Bottleneck Models}
\author{
    Qihan Huang\textsuperscript{\rm 1},
    Jie Song\textsuperscript{\rm 1}\footnote{Corresponding author.},
    Jingwen Hu\textsuperscript{\rm 1},
    Haofei Zhang\textsuperscript{\rm 1},
    Yong Wang\textsuperscript{\rm 2},
    Mingli Song\textsuperscript{\rm 1}
}
\affiliations{
    \textsuperscript{\rm 1} Zhejiang University \\
    \textsuperscript{\rm 2} State Grid Shandong Electric Power Company\\
    \{qh.huang, sjie, jw\_hu, haofeizhang, brooksong\}@zju.edu.cn, wangyong@sd.sgcc.com.cn
}
\usepackage{bibentry}

\begin{document}

\maketitle

\begin{abstract}
Concept Bottleneck Models (CBMs), which break down the reasoning process into the \textit{input-to-concept} mapping and the \textit{concept-to-label} prediction, have garnered significant attention due to their remarkable interpretability achieved by the interpretable concept bottleneck. However, despite the transparency of the concept-to-label prediction, the mapping from the input to the intermediate concept remains a black box, giving rise to concerns about the trustworthiness of the learned concepts~(\textit{i.e.}, these concepts may be predicted based on spurious cues). The issue of concept untrustworthiness greatly hampers the interpretability of CBMs, thereby hindering their further advancement.
To conduct a comprehensive analysis on this issue, in this study we establish a benchmark to assess the trustworthiness of concepts in CBMs. A pioneering metric, referred to as \textit{concept trustworthiness score}, is proposed to gauge whether the concepts are derived from relevant regions. Additionally, an enhanced CBM is introduced, enabling concept predictions to be made specifically from distinct parts of the feature map, thereby facilitating the exploration of their related regions. Besides, we introduce three modules, namely the \textit{cross-layer alignment} (CLA) module, the \textit{cross-image alignment} (CIA) module, and the \textit{prediction alignment} (PA) module, to further enhance the concept trustworthiness within the elaborated CBM.
The experiments on five datasets across ten architectures demonstrate that without using any concept localization annotations during training, our model improves the concept trustworthiness by a large margin, meanwhile achieving superior accuracy to the state-of-the-arts.
Our code is available at \textit{\url{https://github.com/hqhQAQ/ProtoCBM}}.
\end{abstract}

\section{Introduction}

Concept Bottleneck Models~(CBMs) have recently emerged as self-explanatory models that begin by predicting the high-level concepts present in the input data~(\textit{input-to-concept mapping}), and subsequently make class label predictions based on these inferred concepts~(\textit{concept-to-label prediction}). CBMs have garnered substantial attention from researchers due to their inherent interpretability and comparable performance when compared to non-interpretable models.
Following the pioneering work of the first CBM~\cite{Koh2020cbm}, a plethora of CBM variants have emerged in the research landscape. These variants include, but are not limited to, PCBM~\cite{Mert2023post-cbm}, Label-free CBM~\cite{Tuomas2023label-cbm}, Hard CBM~\cite{Marton2022AR-hard}, and Interactive CBM~\cite{Kushal2023interactive-cbm}.

\begin{figure}[t]
\centering
    \includegraphics[width=\linewidth]{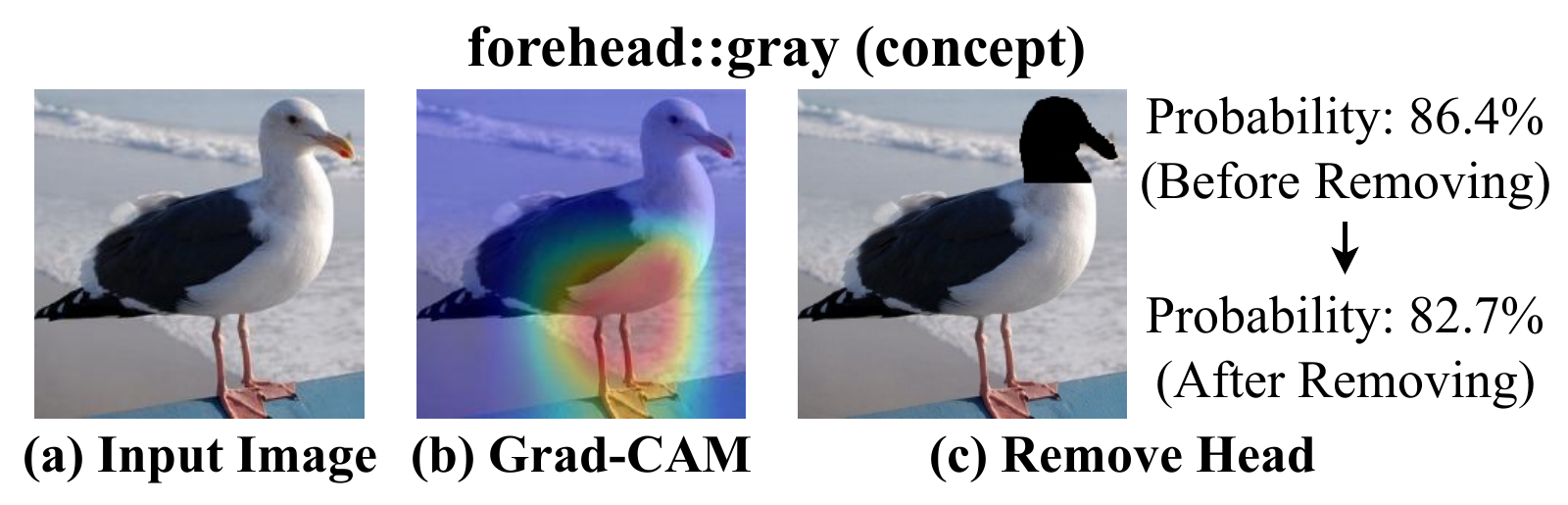}
\caption{Untrustworthiness of a concept named ``forehead::gray'' in the vanilla CBM.
(a) The input image.
(b) The localization map of this concept~(generated by Grad-CAM) concentrates on the underpart of the bird.
(c) After removing the head part in the image, the prediction probability for this concept changes very slightly.}
\label{fig:intro_example}
\end{figure}

Despite the remarkable strides made in the field, the interpretability of existing CBMs primarily stems from the transparency of the concept-to-label prediction. The input-to-concept mapping, in contrast, is still a black box and thus remains elusive, which significantly undermines the trustworthiness of the subsequent concept-to-label prediction.
Figure~\ref{fig:intro_example} illustrates an example highlighting this issue. Specifically, Figure~\ref{fig:intro_example}~(b) demonstrates the attribution map of a head concept prediction using Grad-CAM~\cite{Ramprasaath2017grad-cam}, indicating that the prediction of this head concept is based on the underpart rather than the head part.
Besides, Figure~\ref{fig:intro_example}~(c) further presents that after removing the head part in the input image, the classification probability of this head concept only changes slightly.
While some recent studies have shed light on the concept untrustworthiness in CBMs~\cite{margeloiu2021concept, heidemann2023concept, furby2023end-to-end-cbm}, these works have been limited to either visualization examples or a simple metric applied to a specific CBM on a single backbone. A more comprehensive investigation into concept trustworthiness is urgently needed to drive the progress of CBMs.

In this work, we establish a systematic benchmark on concept trustworthiness with a proposed evaluation metric termed \textit{concept trustworthiness score}.
This metric assesses to what extent the concepts predicted by CBMs are align with the annotated object parts in the dataset. 
To ensure a comprehensive evaluation, our benchmark encompasses various CBM variants implemented across multiple architectures, including a diverse range of CNNs and vision transformers. The experimental results obtained from this benchmark reveal a significant insufficiency in the concept trustworthiness of previous CBMs, thereby highlighting the limitations in their interpretability.

To further enhance the concept trustworthiness, we propose an elaborated CBM framework that deviates from the conventional approach of utilizing average pooling on the final feature map extracted from the backbone network.
The average pooling operation mixes the features of different image regions together and misleads the concept predictions, thus undermining the overall concept trustworthiness.
Therefore, we draw inspiration from the ProtoPNet~\cite{Chen2019ProtoPNet} framework and eliminate the pooling operation altogether. 
In our proposed approach, we define multiple part-prototypes as trainable vectors to represent different object parts. We then iterate over different regions of the feature map to identify and evaluate the presence of these part-prototypes. The activation value of each part-prototype, which signifies the existence of the represented object part in the feature map, is calculated accordingly. Subsequently, a fully-connected layer is employed to make concept predictions based on the activation values of the part-prototypes.

Besides, we propose three modules into the above CBM framework to further improve the concept trustworthiness: a cross-layer alignment~(CLA) module, a cross-image alignment~(CIA) module, and a prediction alignment~(PA) module.
The above CBM framework locates the related image region of each predicted concept in two steps: (1) The prototypes faithfully locate their corresponding object parts in the last feature map; (2) Each concept accurately matches the prototypes that represent its related image regions.
The first step requires that the last feature map is spatially aligned with the input image, and the CLA \& CIA modules are proposed to approach this requirement.
Specifically, the CLA module adopts a multi-scale mechanism to facilitate the spatial alignment between the feature maps in the deep and shallow layers.
The CIA module promotes the spatial alignment between the feature maps of the original image and the augmented image.
Furthermore, to improve the second step, the PA module is proposed to constrain the localization regions~(generated with the matched prototypes) of concepts to be consistent.

We perform comprehensive experiments to validate the performance of our proposed model.
Experiment results demonstrate that without using any concept localization annotations during training, our proposed CBM framework and three modules significantly improve the concept trustworthiness.
Besides, with the decoupled concept learning mechanism that concentrates the learning of concepts, our model achieves state-of-the-art accuracy on five datasets across ten architectures.

To sum up, the key contributions of our work can be listed as follows:

\begin{itemize}
    \item We establish a systematic benchmark to evaluate the concept trustworthiness with a proposed evaluation metric~(concept trustworthiness score) across multiple backbones, unveiling the cons of various CBMs.
    \item We introduce an elaborated CBM model that decouples the feature map for concept prediction, with three proposed modules to further improve the concept trustworthiness: a cross-layer alignment~(CLA) module, a cross-image alignment~(CIA) module, and a prediction alignment~(PA) module.
    \item Experiment results show that our model achieves state-of-the-art performance, in both concept trustworthiness and accuracy, on five datasets across ten architectures.
\end{itemize}

\section{Related Work}

\paragraph{Concept Bottleneck Models.}
Concept bottleneck models~(CBMs) are self-explainable models that make class predictions based on the prior predicted concepts.
With the intermediate human-understandable concepts inside the model, CBMs enable humans to interpret the original black-box model and quickly intervene on the concept prediction for higher accuracy.
After the emergence of the vanilla CBM~\cite{Koh2020cbm}, various variants of CBMs have been proposed.
Label-free CBM utilizes large-scale language model and cross-modal model, GPT-3 and CLIP~\cite{Alec2021CLIP}, to automatically generate concept embeddings, thus reducing the cost concept labeling in CBMs.
PCBM aims to turn any pre-trained deep neural network into a CBM, by leveraging CAVs~(Concept Activation Vectors)~\cite{Kim2019TCAV} to represent concepts and projecting the image features into these learned CAVs for concept prediction.
However, recently some works~\cite{margeloiu2021concept, heidemann2023concept, furby2023end-to-end-cbm} point out that many learned concepts of CBMs are not predicted from the related image regions, thus weakening the interpretability of CBMs.
Therefore, our work aims to establish a systematic benchmark to evaluate the concept trustworthiness of CBMs and propose an enhanced CBM to improve the concept trustworthiness.

\paragraph{Part-Prototype Networks.}
Part-prototype networks are self-explainable models that make class predictions through intermediate interpretable part-prototypes.
Specifically, they define multiple part-prototypes to represent object parts, and mimic humans to make predictions by comparing object parts across different images.
ProtoPNet~\cite{Chen2019ProtoPNet} is the first part-prototype network, with many follow-up part-prototype networks: ProtoTree~\cite{meike2021ProtoTree}, Deformable ProtoPNet~\cite{Jon2022Deform}, TesNet~\cite{jiaqi2021TesNet}, ProtoPFormer~\cite{xue2022protopformer}, ProtoPShare~\cite{Dawid2021ProtoPShare}, and ProtoPool~\cite{Ryma2022Assignment}.
Our work follows part-prototype networks to decouple the feature map and make concept predictions from different parts of the feature map.

\section{Concept Trustworthiness Benchmark}

\subsection{Preliminaries}

The vanilla CBM consists of a feature extractor $f$, a concept predictor $g$, and a category predictor $h$.
Specifically, given an input image $x$ with height $H$ and width $W$, $f$ extracts the feature map $z \in \mathbb{R}^{H_z \times W_z \times D}$~($H_z$, $W_z$, $D$ denote the height, width and dimension of the feature map $z$).
Next, CBM employs average pooling on $z$ to obtain $\bar{z} \in \mathbb{R}^{D}$, then feeds $\bar{z}$ into the concept predictor $g$ to generate the concept classification probabilities $g(\bar{z}) \in \mathbb{R}^{C}$~($C$ denotes the total number of concepts).
Finally, CBM feeds $g(\bar{z})$ into the category predictor $h$ to generate the classification probabilities $h(g(\bar{z})) \in \mathbb{R}^{K}$~($K$ denotes the total number of categories).

CBM adopts two losses for model training: a concept loss $\mathcal{L}_{\mathrm{concept}}(g(\bar{z}), c^{\mathrm{gt}})$ to supervise the concept prediction, and a task loss $\mathcal{L}_{\mathrm{task}}(h(g(\bar{z})), y^{\mathrm{gt}})$ to supervise the category prediction.
Here, $c^{\mathrm{gt}} \in \mathbb{R}^{C}$ denotes the concept label, \textit{i.e.}, $c^{\mathrm{gt}}_{i} = 1$ if the $i$-th concept exists in the image and $c^{\mathrm{gt}}_{i} = 1$ otherwise, and $y^{\mathrm{gt}}$ denotes the category label.
In the later sections, this paper uses $\mathcal{L}_{\mathrm{concept}}$ and $\mathcal{L}_{\mathrm{task}}$ instead of $\mathcal{L}_{\mathrm{concept}}(g(\bar{z}), c^{\mathrm{gt}})$ and $\mathcal{L}_{\mathrm{task}}(h(g(\bar{z})), y^{\mathrm{gt}})$ for simplicity.

\subsection{Concept Trustworthiness Score}

\begin{figure}[t]
\centering
    \includegraphics[width=\linewidth]{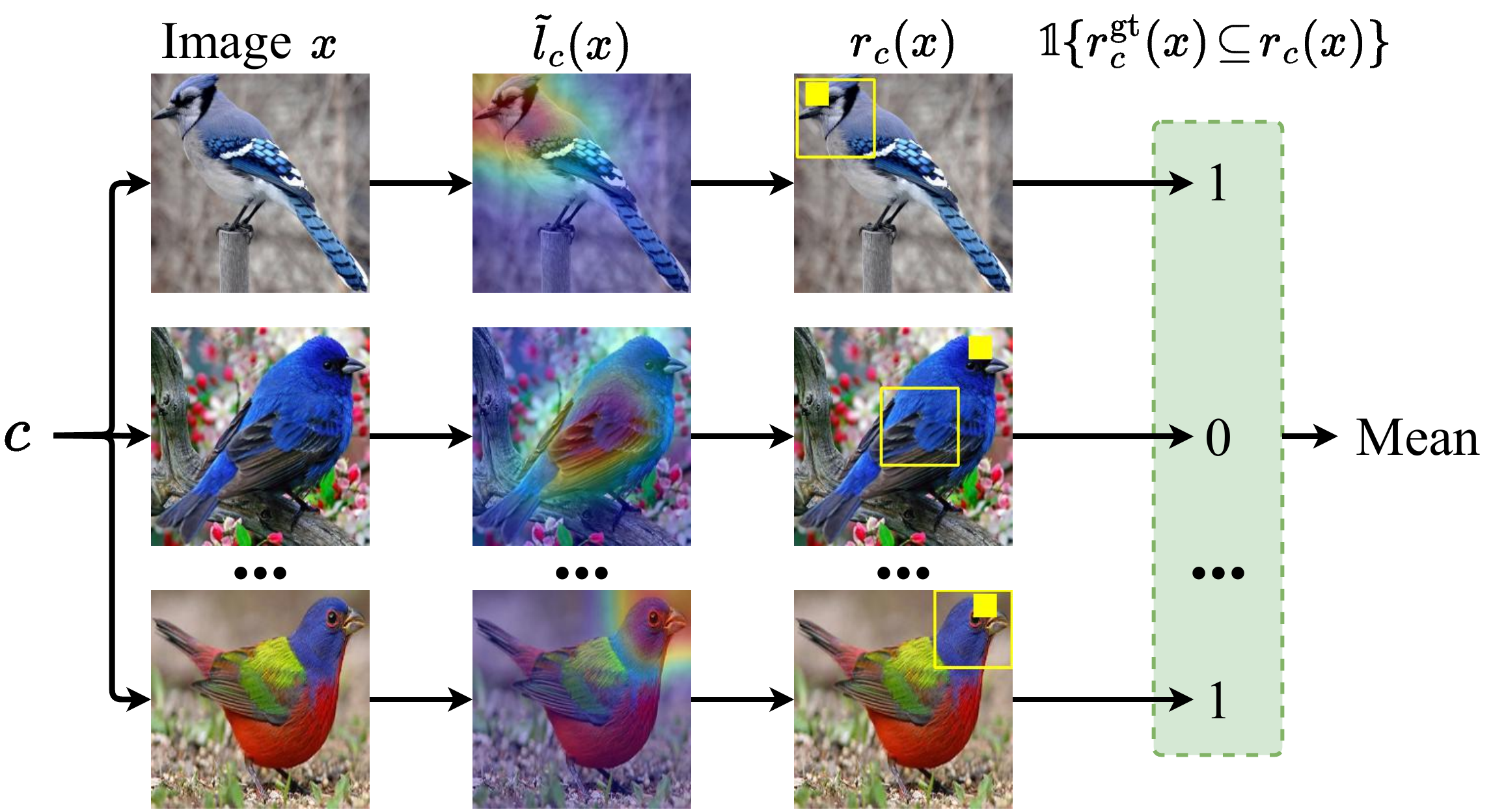}
\caption{The calculation of concept trustworthiness score of a concept $c$ about the forehead of the bird.}
\label{fig:metric_cal}
\end{figure}

Our work establishes a systematic concept trustworthiness benchmark with an evaluation metric named \textit{concept trustworthiness score}.
(\textbf{First step}) To calculate this evaluation metric, our work first generates the corresponding region of each concept on the input image~(\textit{i.e.}, the image region that the concept is predicted from). (\textbf{Second step}) Next, this evaluation metric estimates the concept trustworthiness according to whether the corresponding region of the concept is consistent with the ground-truth~(\textit{i.e.}, the object part annotations in the dataset).

In the first step, our work first calculates the localization map $l_c(x) \in \mathbb{R}^{H_l \times W_l}$ of each concept $c$ on the input image $x$~($H_l$ \& $W_l$ denote the shape of $l_c(x)$, and $l_c(x)$ indicates the important regions for concept prediction in the input image).
For example, for the previous CBMs which lack intrinsic concept localization ability, our work calculates $l_c(x)$ using the attribution methods~(\textit{e.g.}, Grad-CAM~\cite{Ramprasaath2017grad-cam}, Grad-CAM++~\cite{Aditya2018Grad-CAM++}).
Next, our work follows ProtoPNet~\cite{Chen2019ProtoPNet} to resize $l_c(x)$ to be $\tilde{l}_c(x) \in \mathbb{R}^{H \times W}$ with the same shape as $x$, then calculates the corresponding region $r_c(x)$ as a fix-sized bounding box~(with a pre-determined shape $H_b \times W_b$) whose center is the maximum element in $\tilde{l}_c(x)$.

In the second step, our work determines the trustworthiness of concept $c$ according to whether the ground-truth region $r^{\mathrm{gt}}_c(x)$ of concept $c$ is inside $r_c(x)$, which can be described as $\mathbb{1}\{ r^{\mathrm{gt}}_c(x) \! \subseteq \! r_c(x) \}$~($\mathbb{1}\{ \cdot \}$ denotes the indicator function).
Our work estimates the trustworthiness of each concept $c$ on all the images that contain $c$, then calculates the averaged results over total $C$ concepts.
Let $\mathcal{I}_{c}$ denote the images that contain $c$, the concept trustworthiness score $S_{\mathrm{concept}}$ is finally defined as~(note that $\| \cdot \|$ denotes cardinality of a set, and $S_{\mathrm{concept}} \in [0,1]$):

\begin{equation}
    S_{\mathrm{concept}} = \frac{1}{C} \sum\limits_{c=1}^{C} \frac{1}{\| \mathcal{I}_{c} \|} \sum\limits_{x \in \mathcal{I}_{c}} \mathbb{1} \{ r^{\mathrm{gt}}_c(x) \subseteq r_c(x) \}.
\end{equation}

With this proposed evaluation metric, a systematic benchmark can be established for various CBM variants across multiple backbones.
Besides, the proposed benchmark normalizes the object sizes~(\textit{i.e.}, by cropping the objects and resizing them to be the same size), which eliminates the evaluation distractors for a more normative benchmark~(\textit{e.g.}, if the object size is too small in the image, the corresponding regions of different concepts are easily confounded).

\begin{figure*}[t]
\centering
    \includegraphics[width=\linewidth]{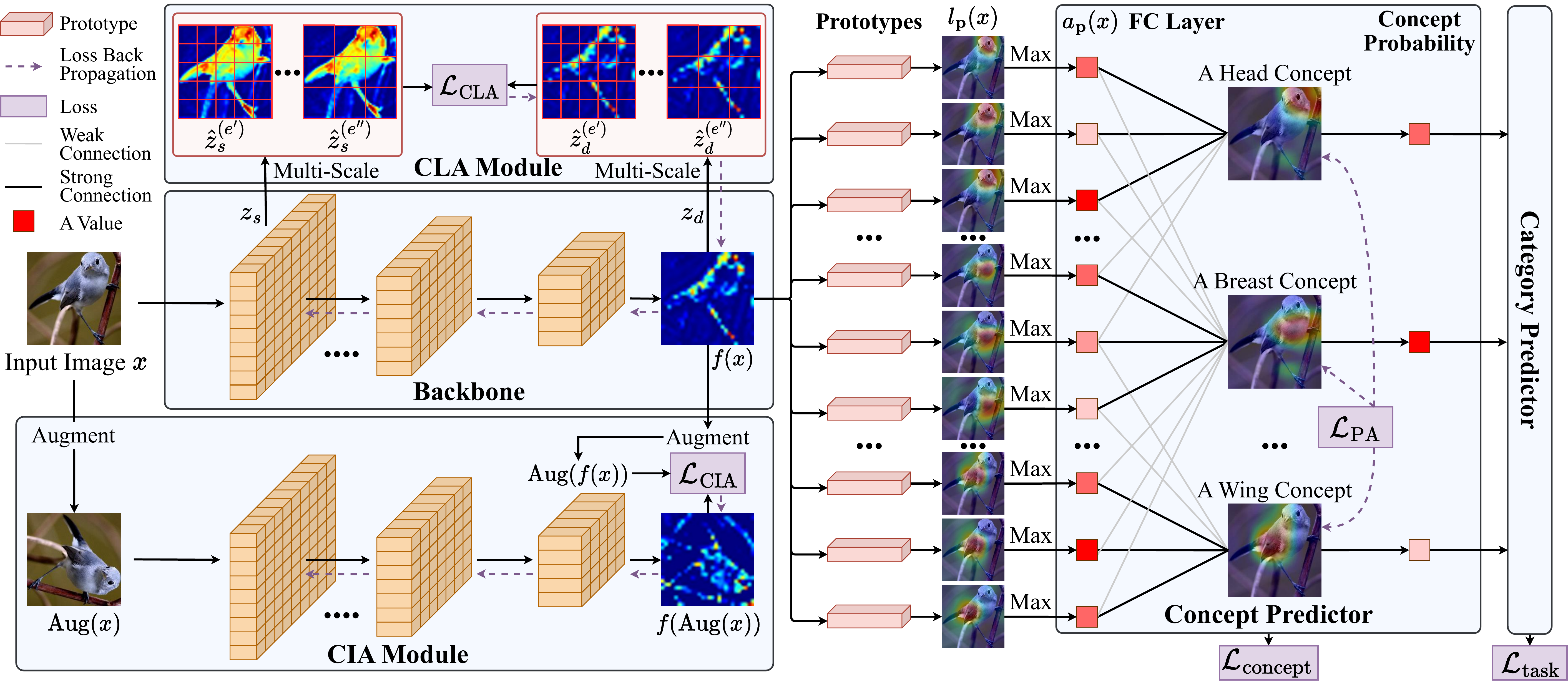}
\caption{Overview of our proposed model~(only three concepts are presented for brevity).
Given the input image $x$, the ``Backbone'' extracts different layers of features for $x$.
The CLA module spatially aligns the last feature map $f(x)$~(also denoted as $z_d$) with the input image according to the shallow feature map $z_s$ in a multi-scale manner.
Meanwhile, the CIA module spatially aligns $f(x)$ with $x$ by aligning $f(x)$ with the feature map of another augmented image.
Next, $f(x)$ is aggregated with multiple prototypes, generating the localization maps $l_{\boldsymbol{\mathrm{p}}}(x)$ and activation values $a_{\boldsymbol{\mathrm{p}}}(x)$ of prototypes.
Finally, the activation values $a_{\boldsymbol{\mathrm{p}}}(x)$ are fed into ``Concept predictor'' and ``Category Predictor'' for concept prediction and category prediction, respectively.
The PA loss is used to supervise the localization maps of the learned concepts.
Note that the loss back propagation of $\mathcal{L}_{\rm concept}$ and $\mathcal{L}_{\rm task}$ is omitted for simplicity in the figure.}
\label{fig:method}
\end{figure*}

\section{Enhanced CBM}

The experiments on this concept trustworthiness benchmark demonstrate that previous CBMs have insufficient concept trustworthiness.
Our work speculates that their concept untrustworthiness results from the average pooling operation on the last feature map.
Specifically, the average pooling operation mixes the features of different image regions together for the further concept prediction, thus easily misleading concepts into learning from their unrelated image regions.
Therefore, our work establishes an elaborated \textbf{CBM framework}, following ProtoPNet~\cite{Chen2019ProtoPNet} to eliminate the pooling operation and make predictions from specific parts of the feature map using the learnable prototypes.
Furthermore, we propose three modules into this framework for higher concept trustworthiness: a \textbf{cross-layer alignment~(CLA) module}, a \textbf{cross-image alignment~(CIA) module}, and a \textbf{prediction alignment~(PA) module}.

\subsection{CBM Framework}

The proposed CBM framework contains $M$ learnable prototypes $\boldsymbol{\mathrm{P}} = \{ \boldsymbol{\mathrm{p}}_j \in \mathbb{R}^{1 \times 1 \times D} \}_{j=1}^{M}$.
Given the feature map $z \in \mathbb{R}^{H_z \times W_z \times D}$ extracted from the input image $x$, our work generates the localization map $l_{\boldsymbol{\mathrm{p}}_j}(x) \in \mathbb{R}^{H_z \times W_z}$ of each prototype $\boldsymbol{\mathrm{p}}_j$ on $z$ by calculating and concatenating the similarity scores between $\boldsymbol{\mathrm{p}}_j$ and each element $\tilde{z}$ of $z$~($z$ consists of $H_z \times W_z$ elements, and the shape of each element is $D$).
Next, the activation value $a_{\boldsymbol{\mathrm{p}}_j}(x)$ of $\boldsymbol{\mathrm{p}}_j$ on $z$ is calculated as the maximum value in $l_{\boldsymbol{\mathrm{p}}_j}(x)$~($\mathrm{Sim}(\cdot,\cdot)$ is the similarity score between two vectors):

\begin{equation}
\begin{aligned}
    a_{\boldsymbol{\mathrm{p}}_j}(x) &= \max \ l_{{\boldsymbol{\mathrm{p}}}_j}(x) \\ &  = \max_{\tilde{z} \; \in \; \mathrm{elements}(z)} \mathrm{Sim}(\tilde{z}, \boldsymbol{\mathrm{p}}_j).
\end{aligned}
\end{equation}

In this manner, each $a_{\boldsymbol{\mathrm{p}}_j}(x)$ is calculated from a specific part of the feature map, thus the learned prototypes can represent the features of the local part instead of the global object.
Therefore, our work makes concept predictions according to these learned prototypes, facilitating the concepts to explore their related prototypes and the corresponding image regions.
Specifically, with the calculated activation values $\{ a_{\boldsymbol{\mathrm{p}}_j}(x) \}_{j=1}^{M}$ of $M$ prototypes, the concept predictor $g$ can be simply implemented as a fully-connected layer with a weight matrix $\omega^g \in \mathbb{R}^{C \times M}$, which takes $\{ a_{\boldsymbol{\mathrm{p}}_j}(x) \}_{j=1}^{M}$ as input.
Finally, the concept classification probabilities $g(\{ a_{\boldsymbol{\mathrm{p}}_j}(x) \}_{j=1}^{M})$ output by the concept predictor are fed into the category predictor to generate the classification probabilities.
This CBM framework is also trained with the concept loss $\mathcal{L}_{\mathrm{concept}}$ and the task loss $\mathcal{L}_{\mathrm{task}}$.

After training, our work calculates the localization map $l_c(x)$ of each concept $c$ by averaging the localization maps of the $N$ most important prototypes for $c$.
Specifically, let $\mathrm{Top}_{N}(\omega_c^g)$ denote the indexes of the $N$ most important prototypes for $c$~(\textit{i.e.}, these $N$ prototypes contribute most positively to the prediction of $c$ in the concept predictor, which is determined by their weights in $\omega_c^g \in \mathbb{R}^{M}$), $l_c(x)$ is calculated as in Equation~\ref{equa:l_c_ours} and it is used for concept trustworthiness evaluation of this CBM framework~($l_c(x) \in \mathbb{R}^{H_z \times W_z}$, note that $H_l = H_z$ here).

\begin{equation}
\label{equa:l_c_ours}
    l_c(x) = \frac{1}{N} \sum\limits_{j \in \mathrm{Top}_{N}(\omega_c^g)} l_{{\boldsymbol{\mathrm{p}}}_j}(x).
\end{equation}


The proposed CBM framework significantly improves the concept trustworthiness by decoupling the feature map for concept prediction, however, it is still restricted.
In this framework, the $l_c(x)$ locates the related image region of $c$ in two steps: (1) The localization map $l_{{\boldsymbol{\mathrm{p}}}_j}(x)$ of each prototype ${\boldsymbol{\mathrm{p}}}_j$ faithfully locates its corresponding object part in the last feature map $z$; (2) The the concept predictor assigns higher positive values to the prototypes in $\omega_c^g$ that represent the related object part for $c$, \textit{i.e.}, $c$ is predicted mainly from these prototypes.
Specifically, the first step requires that the last feature map $z$ is spatially aligned with the input image $x$, \textit{i.e.}, each pixel of the feature map represents the information of the image region with the same position in the input image. However, this requirement is not guaranteed in the deep neural networks, and some works~\cite{hoffmann2021doeslook, huang2022evaluation} doubt the interpretability of ProtoPNet due to this reason. Therefore, our work proposes the cross-layer alignment module and the cross-image alignment module to tackle this problem.
Furthermore, our work proposes the prediction alignment module to improve the second step.

\subsection{Cross-Layer Alignment Module}

To promote the spatial alignment between the last feature map with the input image, the cross-layer alignment~(CLA) module adopts a multi-scale mechanism to align the last feature map $z_d \in \mathbb{R}^{H_d \times W_d \times D_d}$ with the shallow feature map $z_s \in \mathbb{R}^{H_s \times W_s \times D_s}$~(because $z_s$ is closer to the input pixel space).
Specifically, the CLA module aligns the pair-wise element similarities within $z_d$ and $z_s$, according to that intrinsic similarities between elements within the feature maps can be used for comparison~\cite{simon2019cka, huang2022evaluation}.
Let $\hat{z}_d \in \mathbb{R}^{H_d W_d \times D_d}$ and $\hat{z}_s \in \mathbb{R}^{H_d W_d \times \hat{D}_s}$~(note that $\hat{D}_s = D_s \cdot \frac{H_s}{H_d} \cdot \frac{W_s}{W_d}$) denote the resized $z_d$ and $z_s$, then each element $\phi_{i,j}(\hat{z}_d)$ of the pair-wise element similarities $\phi(\hat{z}_d) \in \mathbb{R}^{H_d W_d\times H_d W_d}$ is calculated as $\mathrm{Sim}(\hat{z}_d[i], \hat{z}_d[j])$, and $\phi(\hat{z}_s)$ is calculated likewise.

However, current form of $\hat{z}_d$ and $\hat{z}_s$ only excavates the object similarities at a single scale and neglects necessary object similarities at other scales.
For instance, it is essential to consider both the similarity between the left eye and right eye at a smaller scale, as well as the dissimilarity between the head part and the wing part at a larger scale of a bird object.
Therefore, our work enriches the feature map to be multi-scale to tackle this problem.
Detailedly, let $E$ denote the total levels of scales, then for each level $e \in \{1, 2, ..., E \}$, our work concatenates the feature vectors within every window~(with size $e \times e$) in the original feature map $z$ to generate the feature map $z^{(e)}$.
With the enriched feature maps $\{ \hat{z}_d^{(e)} \}_{e=1}^{E}$ and $\{ \hat{z}_s^{(e)} \}_{e=1}^{E}$, the alignment loss $\mathcal{L}_{\mathrm{CLA}}$ is calculated as~(note that $\| \cdot \|_{2}$ denotes the L2 norm, and $\mathrm{Detach}(\cdot)$ is the detach operation to stop the gradients):

\begin{equation}
    \mathcal{L}_{\mathrm{CLA}} = \frac{1}{E} \sum\limits_{e=1}^{E} \| \phi(\hat{z}_d^{(e)}) - \mathrm{Detach}\big( \phi(\hat{z}_s^{(e)}) \big) \|_{2}^{2}.
\end{equation}

\subsection{Cross-Image Alignment Module}

The cross-image alignment~(CIA) module spatially aligns the last feature map with the input image in a self-supervised manner.
Specifically, given the input image $x$, the CIA module employs spatial augmentation $\mathrm{Aug}$~(\textit{e.g.}, horizontal flip, rotation) on $x$ to generate the augmented image $\mathrm{Aug}(x)$.
Next, the CIA module employs the same spatial augmentation on the feature map $f(x)$ of the original image $x$, and constrains the augmented feature map $\mathrm{Aug}(f(x))$ to be consistent with the feature map $f(\mathrm{Aug}(x))$ of the augmented image.
In this training manner, the feature map is promoted to perceive the spatial information of the images, thus better aligning with the input image.
In practice, the CIA module freezes the feature map $f(x)$ of the original image $x$ to prevent it from slipping into a trivial solution~(\textit{e.g.}, the parameters of feature extractor $f$ become zero).
Therefore, the alignment loss $\mathcal{L}_{\mathrm{CIA}}$ is finally calculated as:

\begin{equation}
    \mathcal{L}_{\mathrm{CIA}} = \| f(\mathrm{Aug}(x)) - \mathrm{Detach}\big( \mathrm{Aug}(f(x)) \big) \|_{2}^{2}.
\end{equation}

\subsection{Prediction Alignment Module}

With the spatially aligned feature map, each concept $c$ is promoted to be predicted from its related image region with the concept loss $\mathcal{L}_{\rm concept}$, because the features of the related object part of $c$ lead to higher prediction accuracy of $c$.
However, some concepts are still predicted wrongly from the unrelated image regions if they are confused with other concepts~(\textit{i.e.}, two concepts are easily confused if they often appear simultaneously in the same images).
Therefore, the prediction alignment~(PA) module adopts a concept grouping loss $\mathcal{L}_{\rm grp}$ and a concept division loss $\mathcal{L}_{\rm div}$ to align the concept prediction.
The concept grouping loss $\mathcal{L}_{\rm grp}$ facilitates the concepts with the same related region to be predicted from the same region, and the concept division loss $\mathcal{L}_{\rm div}$ encourages the concepts with different related regions to be predicted from different regions.

Specifically, let $\{ \mathcal{C}_i \}_{i=1}^{T}$ denote $T$ groups of concepts, where each $\mathcal{C}_i$ contains concepts with the same related region, and $\mathcal{C}_i$ and $\mathcal{C}_j$ have different related regions for $\forall i \neq j$.
Besides, for a localization map $l_c(x) \in \mathbb{R}^{H_z \times W_z}$, the coordinates $k_c(x) \in \mathbb{R}^{2}$ of the center in $l_c(x)$ are calculated as the weighted average of the multiplication of each element in $l_c(x)$ and its coordinates.
Finally, the alignment loss $\mathcal{L}_{\rm PA}$ in the PA module is calculated as below:

\begin{equation}
    \nonumber
    \begin{cases}
        \mathcal{L}_{\rm PA} = \mathcal{L}_{\rm grp} + \mathcal{L}_{\rm div}. \\
        \mathcal{L}_{\rm grp} \! = \! \frac{1}{T} \sum\limits_{i=1}^{T} \sum\limits_{c \in \mathcal{C}_i} \sum\limits_{c' \in \mathcal{C}_i, c' \neq c} \| k_{c}(x) \! - \! k_{c'}(x) \|_{2}^{2}. \\
        \mathcal{L}_{\rm div} \! = \! - \frac{1}{T^2} \! \sum\limits_{i=1}^{T} \sum\limits_{j=1,j \neq i}^{T} \sum\limits_{c \in \mathcal{C}_i} \sum\limits_{c' \in \mathcal{C}_j} \! \| k_{c}(x) \! - \! k_{c'}(x) \|_{2}^{2}.
    \end{cases}
\end{equation}

\subsection{Loss Function}

With the above modules, the total loss function $\mathcal{L}_{\rm total}$ of our proposed CBM model is finally calculated as below:

\begin{equation}
    \mathcal{L}_{\rm total} = \mathcal{L}_{\rm concept} + \mathcal{L}_{\rm task} + \mathcal{L}_{\rm CLA} + \mathcal{L}_{\rm CIA} + \mathcal{L}_{\rm PA}.
\end{equation}

\section{Experiments}

\begin{table*}
\small
\renewcommand\arraystretch{1}
\centering
\setlength{\tabcolsep}{1.45mm}{
\begin{tabular}{c|*2{c}|*2{c}|*2{c}|*2{c}|*2{c}|*2{c}|*2{c}|*2{c}}
  \toprule
\multirow{2}*{\small \textbf{Method}} & \multicolumn{2}{c|}{\small \textbf{ResNet18}} & \multicolumn{2}{c|}{\small \textbf{ResNet34}}  & \multicolumn{2}{c|}{\small \textbf{ResNet152}} & \multicolumn{2}{c|}{\small \textbf{Dense121}} & \multicolumn{2}{c|}{\small \textbf{Dense161}} & \multicolumn{2}{c|}{\small \textbf{DeiT-Ti}} & \multicolumn{2}{c|}{\small \textbf{DeiT-S}} & \multicolumn{2}{c}{\small \textbf{Swin-S}} \\

\cline{2-17}

& \textbf{\small Loc.} & \textbf{\small Acc.} & \textbf{\small Loc.} & \textbf{\small Acc.} & \textbf{\small Loc.} & \textbf{\small Acc.} & \textbf{\small Loc.} & \textbf{Acc.} & \textbf{\small Loc.} & \textbf{\small Acc.} & \textbf{\small Loc.} & \textbf{\small Acc.} & \textbf{\small Loc.} & \textbf{\small Acc.} & \textbf{\small Loc.} & \textbf{\small Acc.} \\

\midrule

{\small Baseline} & N/A & 78.8 & N/A & 82.3 & N/A & 81.5 & N/A & 80.5 & N/A & 82.2 & N/A & 81.2 & N/A & 82.2 & N/A & 83.4 \\
{\small vanilla CBM} & 22.6 & 77.2 & 21.1 & 79.1 & 18.9 & 79.5 & 24.6 & 78.8 & 26.2 & 80.4 & 13.7 & 78.3 & 12.4 & 79.6 & 11.7 & 81.5 \\
{\small Hard CBM} & 26.4 & 76.9 & 25.2 & 80.4 & 20.6 & 81.0 & 28.1 & 77.9 & 20.7 & 78.9 & 15.8 & 79.4 & 12.9 & 80.6 & 10.4 & 82.0 \\
{\small Label-free CBM} & 29.5 & 77.8 & 32.3 & 79.8 & 28.1 & 78.7 & 31.6 & 78.4 & 28.9 & 79.3 & 18.4 & 77.6 & 14.4 & 78.4 & 12.0 & 81.2 \\
{\small PCBM} & 34.1 & 77.3 & 36.9 & 80.5 & 40.1 & 80.8 & 46.4 & 79.3 & 40.5 & 80.7 & 16.2 & 78.6 & 10.7 & 79.1 & 12.7 & 82.2 \\
\hline

{\small \bf Ours} & 44.6 & 77.2 & 43.2 & 80.3 & 47.2 & \textbf{82.2} & 45.9 & \textbf{81.3} & 45.1 & 82.2 & 34.7 & 81.4 & 34.5 & 81.1 & 30.2 & 83.5 \\
{\small \bf Ours + CLA} & 51.2 & 77.8 & 52.4 & 80.0 & 50.4 & 82.1 & 49.6 & 80.8 & 48.6 & 82.4 & 36.5 & 81.2 & 37.4 & 81.0 & 35.3 & 82.8 \\
{\small \bf Ours + CLA + PA} & 57.6 & 77.9 & 64.1 & 80.4 & 64.8 & 81.9 & 65.5 & 80.6 & 63.2 & \textbf{82.7} & 44.7 & 81.0 & 43.1 & 80.8 & 38.9 & 83.5 \\
{\small \bf Ours + CLA + PA + CIA} & \textbf{62.8} & \textbf{78.6} & \textbf{70.5} & \textbf{81.1} & \textbf{67.6} & 82.1 & \textbf{70.2} & 81.0 & \textbf{65.4} & 82.6 & \textbf{47.2} & \textbf{81.4} & \textbf{46.3} & \textbf{81.5} & \textbf{42.4} & \textbf{83.7} \\

\bottomrule
\end{tabular}}
\caption{The comprehensive evaluation of concept trustworthiness and accuracy of CBMs on CUB-200-2011 dataset. The results are over eight backbones pre-trained on ImageNet. Loc. and Acc. denote concept trustworthiness score and accuracy, respectively. Our results are averaged over 4 runs with different seeds. Bold font denotes the best result in CBMs.}
\label{tab:benchmark_main}

\end{table*}

\subsection{Experimental Settings}

\noindent \textbf{Datasets.}
We follow previous CBMs to conduct experiments on CUB-200-2011~\cite{wah2011caltech}, HAM10000~\cite{tschandl2018ham10000}, and CIFAR10 \& CIFAR100~\cite{krizhevsky2009learning}.
In particular, CUB-200-2011 contains concept annotations and location annotations of the object parts for each image, including 312 concepts and 15 object parts that cover the whole body of the bird.
Therefore, our proposed concept trustworthiness benchmark is mainly established on CUB-200-2011.
Specifically, we select the concepts that relate to object parts~(\textit{e.g.}, head shape, wing color) for the concept trustworthiness benchmark.  
Besides, we crop the bird part in the images according to the bounding box annotations in CUB-200-2011, thus eliminating the evaluation distractors to establish a normative benchmark.

\noindent \textbf{Benchmark Setup.}
Our benchmark evaluates five CBMs: vanilla CBM~\cite{Koh2020cbm}, Hard CBM~\cite{Marton2022AR-hard}, Label-free CBM~\cite{Tuomas2023label-cbm}, PCBM~\cite{Mert2023post-cbm}, and ours.
The concept labels are pre-processed following the vanilla CBM.
The results are re-implemented faithfully following their released codes.

\noindent \textbf{Details.}
In the proposed benchmark, $H_b$ and $W_b$ are both set to be 90 for all CBMs.
In the proposed CBM framework, $M$, $D$, $E$, $N$ are set to be 2000, 64, 2, 10.
We train the proposed model for 18 epochs~(5 epochs for warm-up) with Adam optimizer~\cite{Kingma2015adam}, and the learning rate of the model is set to be 1e-4.
In the early stage training, we follow the paradigm of ProtoPNet to train the model.

\begin{table}
\renewcommand\arraystretch{1}
\small
\centering
\setlength{\tabcolsep}{2.2mm}{
\begin{tabular}{c *4{c}}
  \toprule
\textbf{\small Method} & \textbf{\small CUB} & \textbf{\small HAM10000} & \textbf{\small CIFAR10} & \textbf{\small CIFAR100} \\

\midrule

{\small PCBM} & 58.8 & 94.7 & 77.7 & 52.0 \\
{\small \bf Ours} & {\bf 68.3} & {\bf 96.6} & {\bf 81.3} & {\bf 53.8} \\

\cline{1-5}

{\small PCBM-h} & 61.0 & 96.2 & 87.1 & 68.0 \\
{\small \bf Ours-h} & {\bf 69.4} & {\bf 96.4} & {\bf 88.2} & {\bf 69.1} \\

  \bottomrule
  \end{tabular}}
\caption{Accuracy comparison with PCBM \& PCBM-h on four datasets. Bold font denotes the best result.}
\label{tab:benchmark_post_hoc}

\end{table}

\subsection{Concept Trustworthiness Benchmark \label{sec:concept_benchmark}}

With the proposed concept trustworthiness score, we demonstrate the evaluation results of the concept trustworthiness benchmark in 
Table~\ref{tab:benchmark_main}~(over eight backbones pre-trained on ImageNet).
In the table, ``Baseline'' is the simplest non-interpretable with a fully-connected layer on the last feature map for classification.
We select Grad-CAM++~\cite{Aditya2018Grad-CAM++} to generate the localization maps for previous CBMs, because it has the best performance~(the ablation experiments on the choice of CAM methods are in \textbf{S2.1} of the appendix).

As shown in the table, the concept trustworthiness of previous CBMs are quite insufficient to support their interpretability.
For example, the concept trustworthiness score of the vanilla CBM ranges from 11.7 to 26.2, meaning that most of its learned concepts are not trustworthy.
Specifically, each learned concept in the vanilla CBM is not consistent with the human perception~(because it is inferred from its unrelated image regions), which brings serious potential risks to the subsequent concept-based classification.

\subsection{Comparisons with State-of-the-Art Methods}

The performance of our proposed framework and modules is also demonstrated in Table~\ref{tab:benchmark_main}~(``Ours'' denotes the proposed CBM framework based on part-prototypes).
As shown in the table, the CBM framework achieves superior performance to the previous CBMs in both concept trustworthiness score and accuracy, owing to the concept prediction mechanism that acquires prediction clues from specific parts of the feature map.
Next, with the proposed CLA, CIA and PA modules, our model achieves significantly higher concept trustworthiness score~(Averagely it is 40.2 points higher than the vanilla CBM).

Furthermore, we also validate the accuracy of our model on the benchmark proposed by the previous CBM.
Table~\ref{tab:benchmark_post_hoc} demonstrates the performance comparison with PCBM on four datasets: CUB, HAM10000, CIFAR10, and CIFAR100.
Specifically, the backbones for CUB, HAM10000, CIFAR10 \& CIFAR100 are ResNet18, Inception-v3, and CLIP, respectively.
Note that the performance on CUB is significantly lower than the results in Table~\ref{tab:benchmark_main} because the CUB dataset contains only partial training data and the images are not cropped here~(following the setting in the released codes of PCBM).
Compared to PCBM, PCBM-h adopts an auxiliary fully-connected layer on the image features for classification, which can also be utilized in our model~(``Ours-h'').
The comparison results indicate that our model achieves better accuracy than PCBM on these datasets.

\begin{figure}[t]
\centering
    \includegraphics[width=\linewidth]{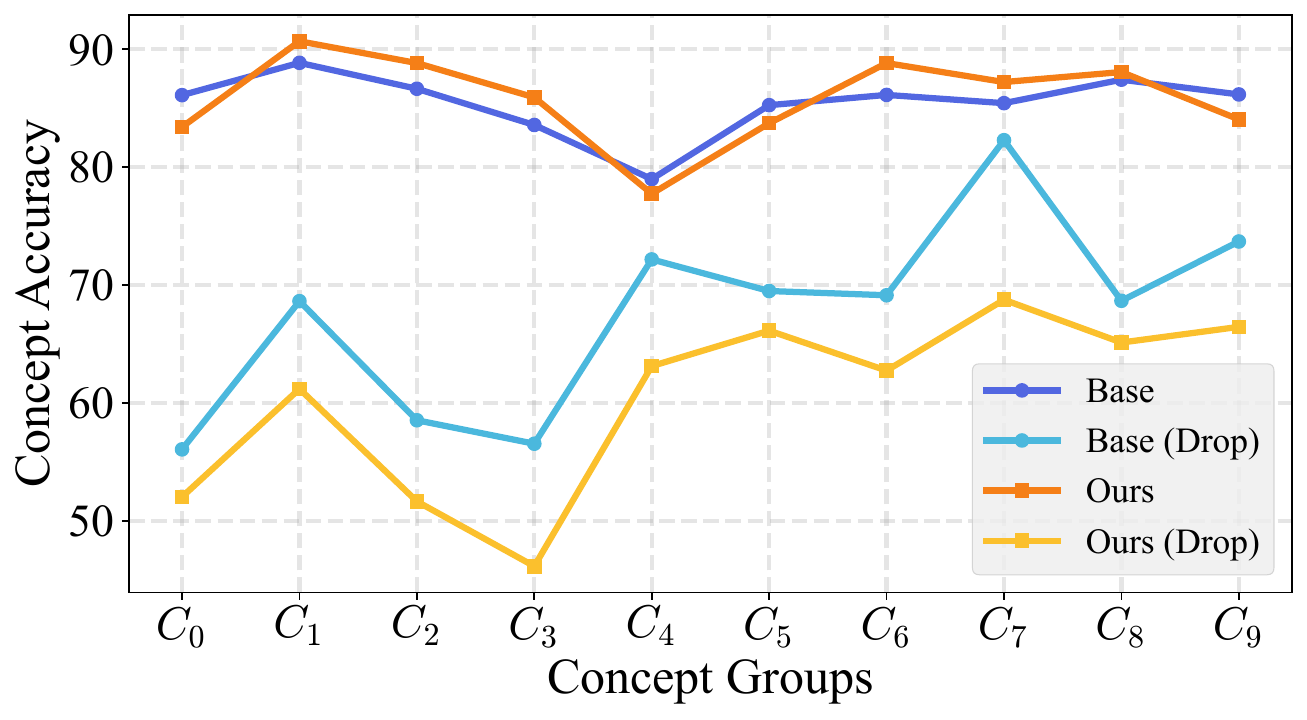}
\caption{The concept prediction accuracy of our model is similar with the base model before dropping the related regions, while it significantly decreases after dropping the related regions~(``Ours~(Drop)'').}
\label{fig:patch_drop}
\end{figure}

\subsection{Ablation Experiments}

Table~\ref{tab:benchmark_main} demonstrates the effectiveness of our proposed CLA, CIA and PA modules.
Specifically, the proposed modules significantly improve the concept trustworthiness score of CBM, while achieving comparable or even better accuracy.
Furthermore, in \textbf{S2.2} \& \textbf{S2.3} of the appendix, we conduct two more ablation experiments to verify that: (1) The CLA \& CIA modules significantly improve the alignment between the deep feature maps and shallow feature maps under the spatial transformation on the original images;
(2) The PA module effectively facilitates the concepts with the same related regions to be predicted from the same region, and encourages the concepts with different related regions to be predicted from different regions.

\paragraph{Patch Drop Experiment.}
We conduct a patch drop experiment to verify that the concepts are rescued from being predicted from the unrelated regions in our proposed model.
Specifically, we evaluate the concept prediction accuracy before/after dropping the related image regions~(by setting the pixels of these regions to be zero, according to the part localization annotations and object segmentation annotations, as described in \textbf{S2.4} of the appendix).
The experiment is conducted over all $T$~($T = 10$) concept groups~(each concept group $\mathcal{C}_i$ contains concepts with the same related region), and the CBM framework without/with the proposed modules~(``Base'', ``Ours'') are evaluated.
As shown in Figure~\ref{fig:patch_drop}, our model has the similar concept prediction accuracy with the base model before dropping the related regions. However, after dropping the related regions, the concept prediction accuracy of our model decreases significantly, indicating that more concepts in our model are predicted from the related regions.

\begin{figure}[t]
\centering
    \includegraphics[width=\linewidth]{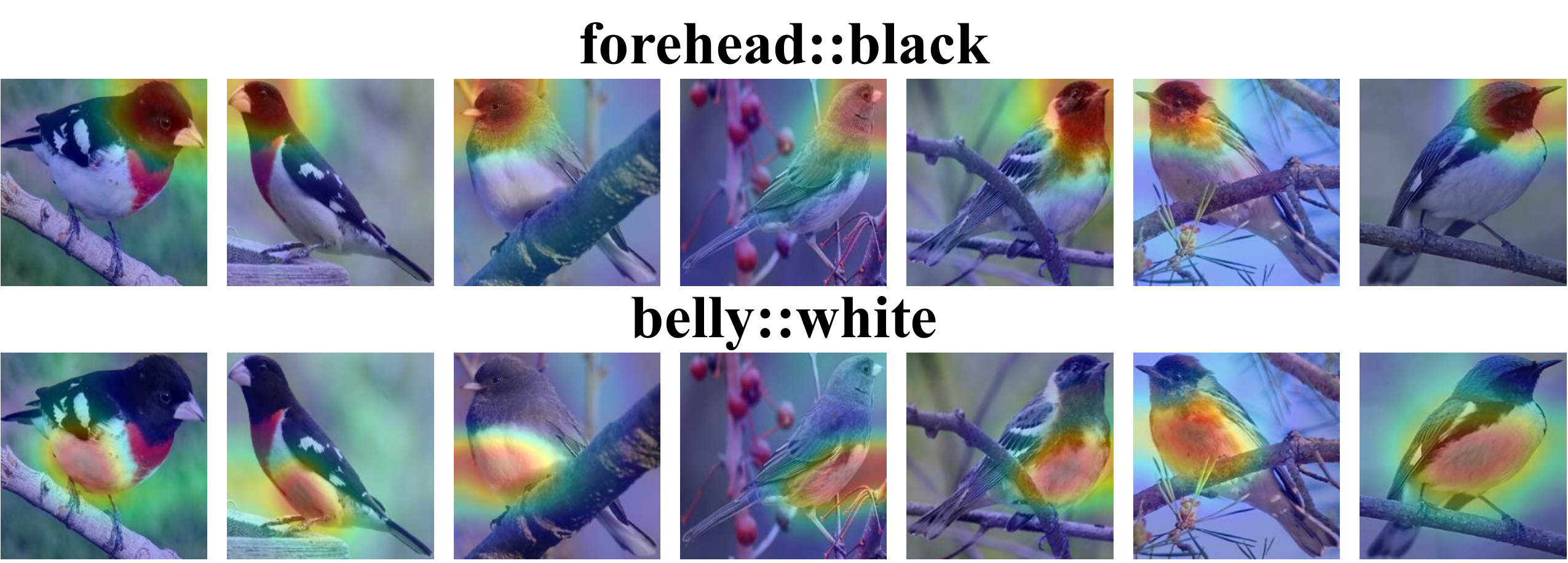}
\caption{The localization maps of two concepts~(``forehead::black'' and ``belly::white'') generated by our model, which accurately locate the related regions of the concepts.}
\label{fig:concept_to_image}
\end{figure}
\begin{figure}[t]
\centering
    \includegraphics[width=\linewidth]{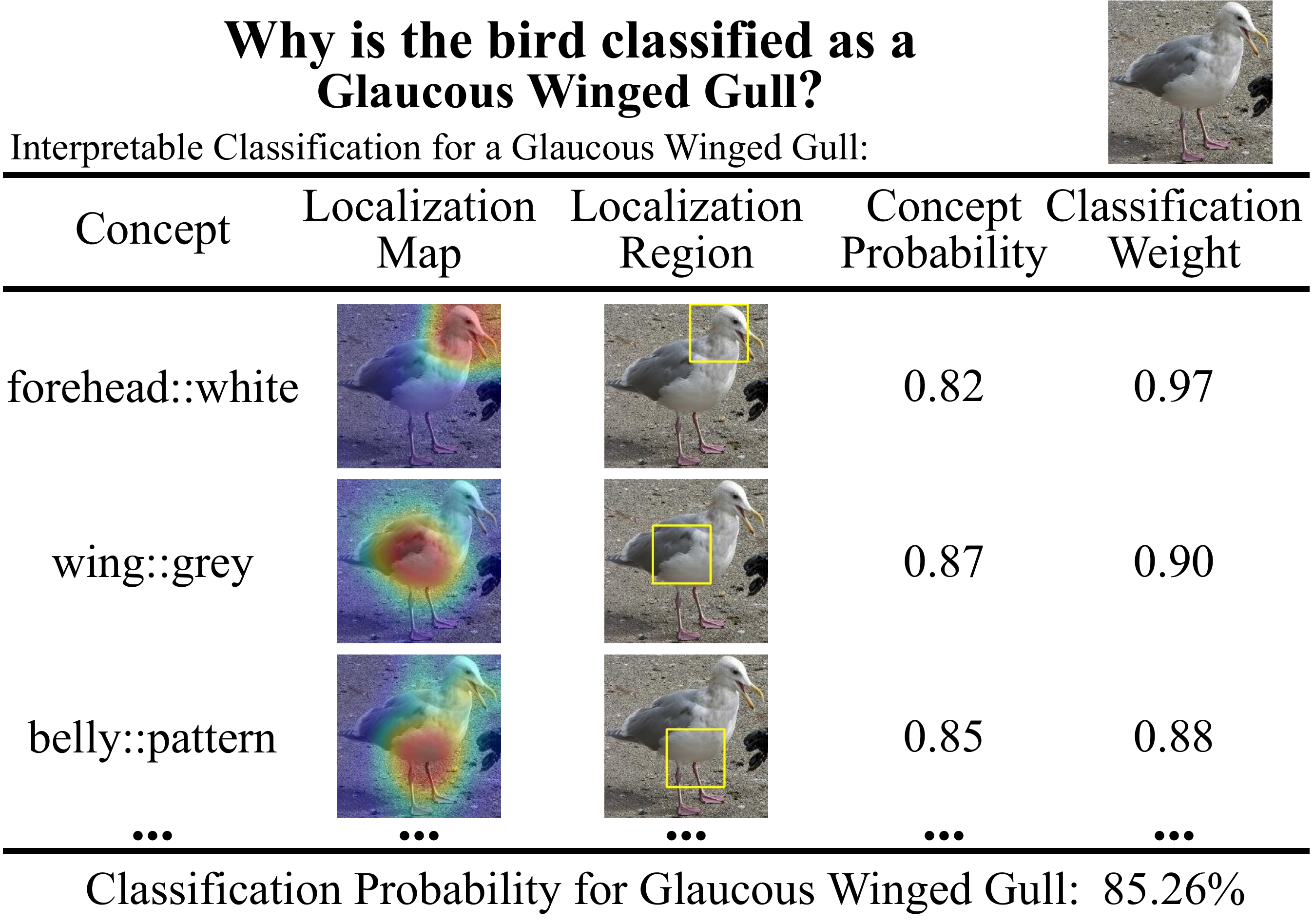}
\caption{Our model first determines the localization map of the concepts, then makes concept predictions on them, and finally predicts the category from the concept probabilities.}
\label{fig:inference_image}
\end{figure}

\subsection{Visualization}

\paragraph{The localization maps of concepts.}
Figure~\ref{fig:concept_to_image} shows the localization maps $l_c(x)$ of two concepts~(``forehead::black'' and ``belly::white'') generated by our proposed model, indicating that our proposed model accurately makes concept predictions from their related image regions.

\paragraph{Interpretable classification of the image.}
Figure~\ref{fig:inference_image} demonstrates the interpretable classification of a Glaucous Winged Gull in our proposed model.
Specifically, our model first determines the localization map of a group of concepts, then predicts the concept probabilities according to the localization maps.
Finally, the classification probability of this bird is calculated by aggregating the predicted concept probabilities through a fully-connected layer.
In this figure, three concepts~(``forehead::white'', ``wing::grey'', ``belly::pattern'') with the highest classification weights~(in the fully-connected layer) are listed.

We also provide more visualization analysis on our proposed model in \textbf{S3} of the appendix for a comprehensive understanding of our method.

\section{Conclusion}

Our work aims to address the concept untrustworthiness problem in concept bottleneck models~(CBMs).
First, we establish a concept trustworthiness benchmark to systematically evaluate current CBMs, based on an evaluation metric named \textit{concept trustworthiness score}.
Next, we propose a CBM framework that utilizes part-prototypes to make concept predictions from specific parts of the feature maps.
Furthermore, we propose three modules~(CLA module, CIA module, and PA module) into this CBM framework to improve the concept trustworthiness.
The comprehensive experiments validate that our model achieves the state-of-the-art performance compared to previous CBMs, in both concept trustworthiness and accuracy.
On the whole, our work conducts an in-depth analysis on the concept untrustworthiness problem in CBMs, towards reducing the potential risks of CBMs in the future.

\section{Acknowledgements}

This work is supported by the Science and Technology Project of SGCC: Research and Digital Application of High-precision Electric Power Super-scale Pre-trained Visual Model (5108-202218280A-2-395-XG).

\bibliography{aaai24}

\end{document}